\documentclass[]{IEEEapm}
\usepackage[utf8]{inputenc}
\usepackage[T1]{fontenc}
\usepackage{graphicx}
\usepackage{url}
\usepackage{tabularx,array}
\jmonth{April}
\pubyear{2027}
\begin{document}
\title{Antenna Design at a Turning Point\\ Giving general-purpose AI agents the freedom to choose their tools}
\author{Dominik~Mair,~\IEEEmembership{Member,~IEEE}, and Daniel~Baumgarten,~\IEEEmembership{Member,~IEEE}}
\affil{Department of Mechatronics, University of Innsbruck, 6020 Innsbruck, Austria}
\maketitle
{\renewcommand{\thefootnote}{}\footnotetext{This work has been submitted to the IEEE for possible publication. Copyright may be transferred without notice, after which this version may no longer be accessible.}\addtocounter{footnote}{-1}}
\markboth{IEEE Antennas and Propagation Magazine}{Industry Activities}
For decades, antenna design has asked how to arrive at a suitable structure: start from a known topology and tune or modify it, guided by experience and analysis, or let an algorithm, from evolutionary search to topology optimization, generate the structure. In November 2026, a further question arises: who chooses this route? We can already ask a general-purpose AI agent to carry out the complete design: build a model, run a full-wave solver and revise the design. Such an agent is a large language model, in our case Claude in Claude Code, that does not only answer in text but calls software tools in a loop: it starts a simulation, reads the result and decides the next step. Anyone whose solver offers its functions to such an agent through an MCP (Model Context Protocol) interface can try this today. More precisely, we must decide which parts of that process the agent should perform itself, and which it should delegate to tools that already contain our engineering knowledge. A standard patch, a microwave filter and an unfamiliar wearable antenna do not require the same design strategy.

Our aim is a verified design with an appropriate amount of work. Below we describe where we stand, what is available and where we think the approach should go. A change is already visible: useful design autonomy depends on giving an agent access to suitable methods and allowing it to choose among them. That changes how we think about automation, specialized AI models and topology optimization.

\section*{Building on the antenna agent}
Cui et al. recently described an LLM-assisted antenna agent on the Cadence Sigrity platform ~\cite{r1}. Their system coordinates literature extraction, topology synthesis, full-wave simulation and Bayesian optimization through MCP. Its examples range from reproducing and rebanding a patch antenna to synthesizing wearable antennas. The work demonstrates the value of treating the design process itself as a task for an agent. Their agent also chooses between capabilities. For a new design from a specification, however, the route is set in advance: a tree search generates the topology and a Bayesian optimizer tunes it, with about 800 full-wave simulations per case.

Our approach uses Hertzon ~\cite{r2}, an open-source modelling and simulation environment built around the EMerge finite-element solver ~\cite{r3}. Hertzon offers geometry operations, simulations, results, design assistants and numerical optimization through MCP. A general-purpose coding agent can use these capabilities without being trained specifically for the benchmark tasks considered here.

\section*{Let the agent choose}
An engineer does not solve every design problem with one fixed algorithm. For a familiar structure, the engineer may start with a textbook model and tune one or more design parameters. For an unfamiliar structure, the engineer may explore several geometries before optimizing any of them. An AI agent should have the same choice. Its first useful decision may be recognizing that the requested design already has a dedicated tool.

\begin{figure}[!t]
\centering
\includegraphics[width=\textwidth]{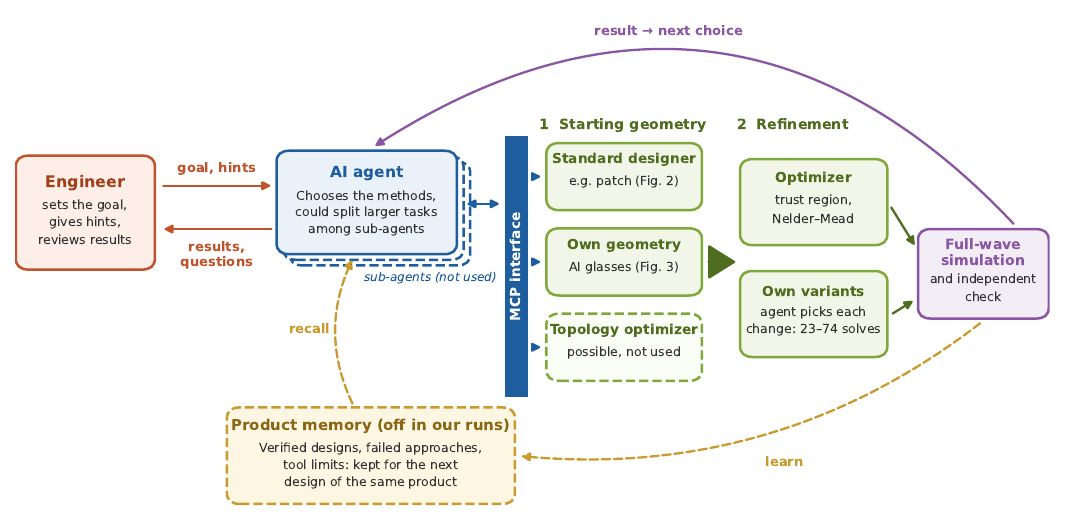}
\caption{Through the MCP interface the agent chooses how to obtain a starting geometry (a standard designer, its own geometry or, in principle, a topology optimizer) and how to refine it (a numerical optimizer or variants it chooses itself). The patch took the upper route, and the AI-glasses designs used own geometries and own variants. The engineer sets the goal and stays in the loop: the agent reports results and questions, and hints such as the one for design A enter as hypotheses that the agent tests. One agent sufficed for our tasks. Dashed elements were not used here: sub-agents for larger tasks, a topology optimizer, and a product memory that would keep what the agent learned for later designs of the same product.}
\label{fig1}
\end{figure}

Fig.~\ref{fig1} shows this setup. The agent receives the design request and calls Hertzon's tools through the MCP interface: a standard designer or its own geometry provides a starting point, an optimizer or its own variants refine it, and each simulation result informs the next choice. Selecting a standard designer, dividing a task into subproblems and proposing a new geometry should all be available choices.

Fig.~\ref{fig1} also contains a memory, deliberately kept empty in our runs so that they could be compared. In product development, an agent working on the same product over several design cycles should keep verified designs, failed approaches and tool limits, but not notes that merely work around tool defects.

\section*{A standard patch should be a simple task}
Consider an inset-fed rectangular patch at 2.45\,GHz. Hertzon provides a dedicated patch designer, available as create\_patch\_antenna. We gave a fresh agent session, without memory, only the request to design a 2.45\,GHz patch antenna on 1.6\,mm FR4 with $|S_{11}|$ $\le$ $-$20\,dB and to verify it in full wave. The prompt named no tool. The agent read the tool descriptions of the Hertzon MCP server and chose create\_patch\_antenna with its built-in trust-region optimization, because the tool is made for patches and the optimizer needs few full-wave solves. This is the intended route: a tool describes itself, and the agent decides whether it fits. The designer creates the parameterized geometry, materials, port, analysis setup and optimization problem, with starting dimensions from the textbook transmission-line model.

\begin{figure}[!t]
\centering
\includegraphics[width=\textwidth]{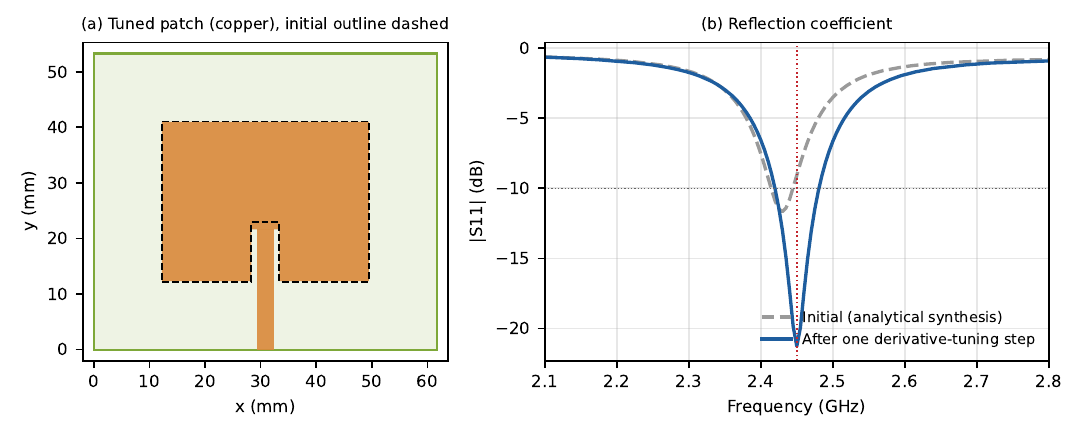}
\caption{A 2.45\,GHz inset-fed patch on FR4 designed by an agent that chose create\_patch\_antenna itself. (a) Tuned geometry, starting design dashed. (b) $|S_{11}|$: one tuning step moves the minimum from 2.43\,GHz to 2.45\,GHz ($-$21.3\,dB).}
\label{fig2}
\end{figure}

For our 2.45\,GHz patch on 1.6\,mm FR4, with the goal $|S_{11}|$ $\le$ $-$20\,dB at 2.45\,GHz, two designs were solved in full wave (Fig.~\ref{fig2}). The analytical starting design resonated at 2.43\,GHz and reached only $-$9.1\,dB at 2.45\,GHz. One derivative-tuning step changed the patch length from 28.81 to 28.64\,mm and the inset depth from 10.69 to 9.33\,mm. An independent full-wave check of this geometry gave $-$21.3\,dB with the minimum at 2.45\,GHz. In this step, Hertzon moves the mesh nodes with the design parameters, estimates the correction from a reduced model and checks it with one full-wave solve. Such tuning refines an existing topology but adds no new branch or slot. The optimization took 93\,seconds, and the whole session, from the request to the verified design, took about six minutes.

\section*{When the topology is not already available}
The industrial AI-glasses challenges introduced by Zhu et al. ~\cite{r4} require a different starting point. Both ask for high total efficiency in the 2.4\,GHz and 5–7\,GHz Wi-Fi bands under fixed mechanical constraints. One provides a 25 $\times$ 5\,mm design window, the other a flexible circuit of roughly 15 $\times$ 3\,mm near the main board. The surrounding structure is prescribed.

Here, the agent must explore geometries, and its decisions matter more than any single number. In the first benchmark it started with an inverted-F antenna, added a grounded parasitic element and, when that combination still needed a matching network, rebuilt the antenna as a coupled-fed shorted strip that covered both bands without one. In the second benchmark, it added a grounded parasitic element beside the feed of design A, which moved a resonance from above 8\,GHz to about 7\,GHz. For design C, a diagnostic simulation showed that its upper-band branch only loaded the third harmonic of the low-band arm instead of creating a second resonance. The agent abandoned the branch and moved the short and feed to the middle, with the low-band arm on one side and the upper-band arm on the other. This T shape produced two upper-band resonances.

None of these final designs came from derivative tuning or another optimizer. The dimensions of the first benchmark design and of A were set by parametric studies, those of B and C by individual simulations that the agent chose one at a time.

\section*{Results of the two benchmarks}
For the first benchmark, the agent found the coupled-fed antenna in approximately two and a half hours and 50 simulations on a workstation. An independent commercial finite-element solver, Ansys HFSS, using the original benchmark CAD, gave mean total efficiencies of about 85\,\% in the 2.4\,GHz band and 77\,\% in the 5–7\,GHz band. No matching components were used.

\begin{figure}[!t]
\centering
\includegraphics[width=\textwidth]{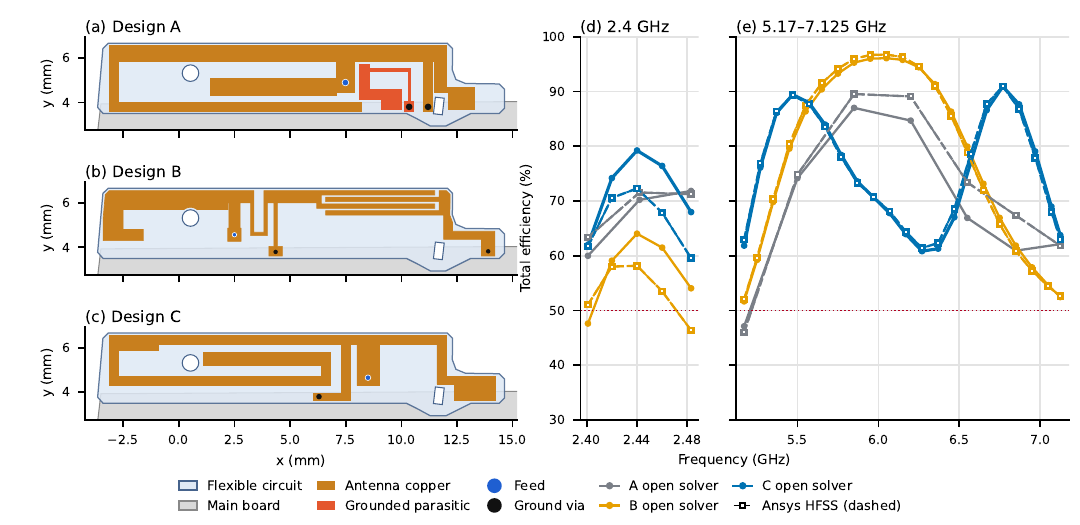}
\caption{Second benchmark. (a)–(c) Designs A, B and C on the flexible circuit (benchmark geometry from ~\cite{r4}), all without matching components. (d), (e) Their total efficiency in the open solver (solid) and in Ansys HFSS (dashed). The dotted line marks 50\,\%.}
\label{fig3}
\end{figure}

The smaller second benchmark produced several structures (Fig.~\ref{fig3}\,(a)–(c)). Table~\ref{tab1} collects the Ansys HFSS results for designs A, B and C. Design C is the only one of these designs that exceeds 50\,\% at every evaluation frequency in both solvers (Fig.~\ref{fig3}\,(d), (e)).

\begin{table}[!t]
\centering
\caption{Second benchmark: goals, design effort without verification, and total efficiency in Ansys HFSS. No matching components}
\label{tab1}
\footnotesize\renewcommand{\arraystretch}{1.3}
\begin{tabularx}{\textwidth}{@{}l >{\raggedright\arraybackslash}X l l l@{}}
\hline
\textbf{Design} & \textbf{Goal} & \textbf{Time, simulations} & \textbf{2.4\,GHz mean (min)} & \textbf{5–7\,GHz mean (min)} \\
\hline
A & simple instruction, then extra 6–7\,GHz resonance & 6.5\,h, 74 (3 runs) & 68.7\,\% (63.4\,\%) & 71.7\,\% (46.0\,\%) \\
B & first $\ge$ 40\,\% at every frequency, then $\ge$ 50\,\% & 5.5\,h, 39 + 2\,h, 7 & 53.4\,\% (46.4\,\%) & 77.8\,\% (52.1\,\%) \\
C & $\ge$ 50\,\% at every frequency from the start & 2.5\,h, 23 & 66.4\,\% (59.6\,\%) & 75.6\,\% (61.4\,\%) \\
\hline
\end{tabularx}
\end{table}

\section*{Break the task into solvable parts}
The same principle applies beyond antennas. Hertzon also provides design assistants for structures such as hairpin filters and diplexers, together with numerical optimization methods. A filter designer can encode synthesis knowledge, generate a useful starting model and expose a smaller tuning problem. The agent need not rediscover that knowledge independently for every request. A two-simulation outcome for the patch should not, however, be assumed for every filter or other structure.

A larger task may be easier when divided into parts that have well-understood solutions. The agent can identify those parts, select appropriate methods and use simulation to check whether their combination meets the overall requirements. Decomposition is a design decision: the interactions between parts still need to be evaluated in the assembled model.

This also opens a route to larger problems. For a patch array with a feed network, an orchestrating agent could split the task into the radiating element, filters, a power divider and phase-shifting lines, give each to a fast sub-agent using a standard designer, and check the assembled array (Fig.~\ref{fig1}). We have not yet demonstrated this.

This is familiar engineering practice. Making it available to an agent means providing tools with clear inputs, meaningful outputs and explicit limits. A low-level geometry interface remains useful, but a validated synthesis tool or efficient local optimizer can remove much more work from a standard task. The freedom to choose includes the freedom to use the simplest method.

\section*{What the experiments do and do not show}
Design A was developed with Claude Opus 5.5 in three consecutive runs from by far the simplest instruction, with an expert hint between runs (see below). Designs B and C were designed with the more recent Claude Fable 5.2 from the benchmark files, with A's efficiency curves as the result to beat. Hertzon was updated between runs. B received two goals in sequence (Table~\ref{tab1}), C the second goal from the start. These different histories prevent a controlled ranking based on runtime or simulation count. They also explain part of the B–C difference: for 50\,\%, B only reshaped its topology, although its report called for a second upper-band resonance, whereas C sought one from the start. Repeated runs would show how much is chance.

We deliberately did not give the agent a narrow framework. No fixed sequence of topology search and numerical optimization was prescribed, and using Hertzon's optimizers was the agent's choice. For B and C it set up none, although their instructions asked it to keep refining until further simulations brought no gain. A topology optimizer could be offered in the same way. Engineers turn to one when familiar geometries fail, and an agent can make the same decision. Early experience suggests that a tight harness of prompts and rules can act like a fetter: the harder the agent is pushed in one direction, the less creative it becomes and the longer it may take. How much freedom a harness should give deserves research.

Cui et al.'s wearable example ~\cite{r1} was evaluated beside a head model, ours in free space, so our figures do not establish a performance advantage. Both works instead show the value of agents coordinating engineering tools.

Verification remains essential. In an earlier run, solver defects led to misleading results and the agent supplied a plausible explanation for them. An independent solver exposed the discrepancy. Freedom in choosing a design does not remove the need for convergence and independent checks.

\section*{Intuition stays and the agent tests it}
Engineering intuition remains important. Between the runs for design A, the Chair of the IEEE AP-S Industrial Initiative Committee remarked that the losses near 6.5\,GHz were not caused by mismatch and asked whether a further resonance at 6–7\,GHz could help. The agent treated this as a hypothesis: a lossless model changed the efficiency there by less than one point, and the currents showed no radiating mode. It then added the grounded parasitic element. An engineer might next ask whether the antenna position should change, and the agent could compare several placements. The development is therefore not toward an agent that is simply told to do the work, but toward an assistant that forms hypotheses with the engineer and tests them.

\section*{A changing role for topology optimization}
Part of our motivation for developing topology-optimization methods was to make design knowledge useful outside a specialist laboratory and to relieve engineers of repetitive design work. One route is topology optimization: evolutionary optimization of pixelated antennas, as in our earlier work ~\cite{r5}, density-based methods such as SIMP ~\cite{r6,r7}, level-set methods ~\cite{r8}, or combinations of these. General-purpose agents offer another route: they select existing knowledge, use it through tools and revise a geometry when needed.

A new optimization method should therefore be judged by one question: Does it provide a capability or a measurable advantage that an agent with access to established synthesis, local tuning and flexible geometry operations cannot already obtain? The comparison should include total computational effort, reliability across repeated runs and the quality of independently verified designs.

Specialized training should face the same question. A general-purpose model with suitable tools should be tested before concluding that a new task requires a dedicated, antenna-specific model.

\section*{A fixed workflow or an organizing agent?}
An agent that changes and evaluates geometries performs part of the topology search itself. Our examples therefore do not indicate that topology optimization loses its importance. It may remain valuable for design spaces that familiar geometries do not cover, as one tool among several that an agent selects.

For now, the practical direction is clear: give agents access to the engineering tools we would want for ourselves, let them select and combine methods, and judge the verified result together with the effort required to obtain it.

Do we still need a purpose-built, fixed optimization workflow for every design task, or will a general agent increasingly organize that workflow itself?

\section*{Acknowledgment}
The authors thank Jiang Zhu, Chair of the Industrial Initiative Committee of the IEEE Antennas and Propagation Society, for the benchmark initiative and for helpful comments on this column.

Use of AI: the antenna designs were produced by AI agents (Claude Opus 5.5 and Claude Fable 5.2, Anthropic, in Claude Code). Claude also assisted in editing all sections and in drafting parts of the text and wrote the scripts for Figures 1–3. The authors checked all content and take full responsibility.

\section*{Author Information}
\textbf{Dominik Mair} (dominik.mair@uibk.ac.at) is with the Department of Mechatronics, University of Innsbruck, 6020 Innsbruck, Austria.

\textbf{Daniel Baumgarten} (daniel.baumgarten@uibk.ac.at) is with the Department of Mechatronics, University of Innsbruck, 6020 Innsbruck, Austria.
\end{document}